\documentclass[runningheads]{llncs}
\usepackage[T1]{fontenc}
\usepackage{graphicx,verbatim}
\usepackage{amssymb}
\usepackage{amsmath}
\usepackage{color}
\usepackage{hyperref}
\usepackage{mathrsfs}
\usepackage{multicol, multirow}
\usepackage{xcolor}
\usepackage{makecell, tabularx}
\usepackage{subcaption}
\usepackage{multirow}
\usepackage{booktabs}

\definecolor{darkgreen}{RGB}{0,100,0}

\DeclareMathOperator*{\argmax}{argmax}

\newcommand{\myparagraph}[1]{\smallskip\noindent\textbf{#1.}}

\begin{document}

\title{IP-CRR: Information Pursuit for Interpretable Classification of Chest Radiology Reports}

\titlerunning{IP-CRR}

\author{Yuyan Ge\inst{1}\thanks{Corresponding author. Email: yyge@seas.upenn.edu}  \and
Kwan Ho Ryan Chan\inst{1} \and
Pablo Messina\inst{2} \and
Ren\'e Vidal\inst{1}}
%
\authorrunning{Y. Ge et al.}
%
\institute{University of Pennsylvania, Philadelphia, USA \\
 \and
Pontifical Catholic University of Chile, CENIA, iHEALTH, Santiago, Chile \\
}

\maketitle             
\begin{abstract}
The development of AI-based methods to analyze radiology reports could lead to significant advances in medical diagnosis, from improving diagnostic accuracy to enhancing efficiency and reducing workload. However, the lack of interpretability of AI-based methods could hinder their adoption in clinical settings. In this paper, we propose an interpretable-by-design framework for classifying chest radiology reports. First, we extract a set of representative facts from a large set of reports. Then, given a new report, we query whether a small subset of the representative facts is entailed by the report, and predict a diagnosis based on the selected subset of query-answer pairs. The explanation for a prediction is, by construction, the set of selected queries and answers. We use the Information Pursuit framework to select the most informative queries, a natural language inference model to determine if a fact is entailed by the report, and a classifier to predict the disease. Experiments on the MIMIC-CXR dataset demonstrate the effectiveness of the proposed method, highlighting its potential to enhance trust and usability in medical AI. Code is available at: \url{https://github.com/Glourier/MICCAI2025-IP-CRR}.


\end{abstract}

\setcounter{footnote}{0}
\section{Introduction}
When a doctor makes a decision that can potentially have life-changing consequences for a patient, the transparency of the decision-making is invaluable. As machine learning (ML) methods become more integrated into healthcare, there is a pressing need for safe and trustworthy models. For a prediction task, interpretability serves as a crucial aspect of model transparency that answers ``why'' and ``how'' a model made a prediction, fostering user trust. 

Traditionally, interpretability is carried out by first training a black-box model that optimizes performance, and then producing an explanation in the form of heatmaps~\cite{2017gradcam,2018gradcampp}, selection of features~\cite{2016why,2017shap}, or more generally, attributions~\cite{2017axiomatic,2017feature}. These post-hoc explanation methods are widely used in many areas such as computer vision, finance, and healthcare. However, these methods lack faithfulness~\cite{2019stop}, because the explanation could potentially be an artifact of the explanation method rather than evidence of the model's underlying reasoning. To address this, interpretable-by-design models were proposed to make predictions solely based on a set of interpretable factors~\cite{2018activation}, which usually take the form of a set of concepts~\cite{2020cbm} or queries~\cite{2022ip} about the input. For example, disease diagnosis for a patient can rely on a list of symptoms, which naturally explain the final prediction. While these methods improve faithfulness, they often require making careful and meticulous designs about the model architecture, optimization, and the set of interpretable concepts or questions. This complexity poses a significant challenge, particularly in the medical domain, where the need for both accuracy and interpretability are critical.

\begin{figure}[t]
    \centering
    \includegraphics[width=0.97\textwidth]{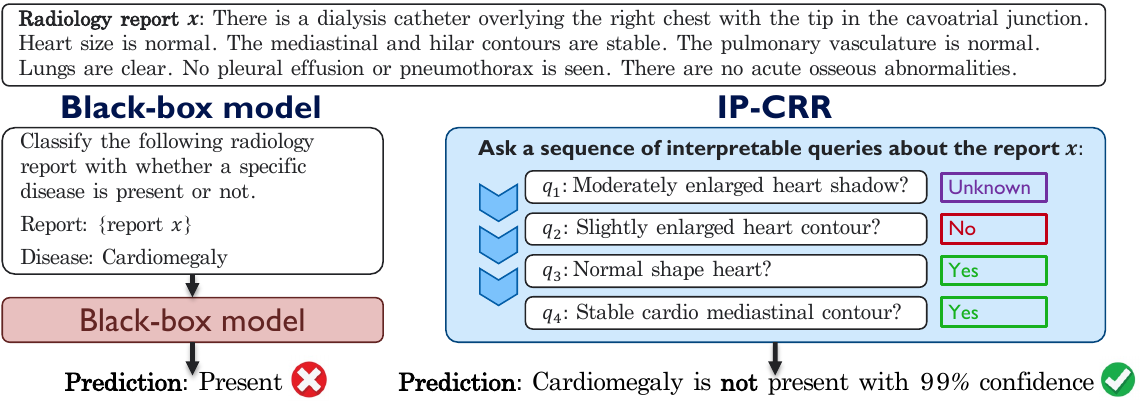}
    \caption{Illustration of the interpretable-by-design method.}
    \label{fig:example_intro}
\end{figure}

Take the task of Chest Radiology Report (CRR) classification as an example (Figure~\ref{fig:example_intro}). A report typically contains a few key insights about the patient, and the task is to predict the associated diseases. For predictions to be interpretable, they must be grounded on a set of queries about the report or patient, such as ``\textit{Is the heart contour enlarged?}'' or ``\textit{Does the lung volume decrease?}''. This requires both the queries and their corresponding answers. Importantly, some queries may lack readily available answers in the report. This setting drastically differs from most settings in interpretable-by-design methods, where all answers are typically available and easily obtained using out-of-the-shelf models~\cite{2024fmvip}.

In this work, we propose an interpretable-by-design framework for CRR classification, which extends the Information Pursuit (IP) framework from image domain~\cite{2022ip,2023vip} to text. Our proposed IP-CRR method involves: 1) curating a set of interpretable queries by mining them from existing CRRs; 2) applying natural language inference (NLI) to determine if a fact is entailed by the report; and 3) learning a classifier that makes predictions by sequentially selecting a list of queries in order of information gain, answering those queries, and predicting the class from the sequence of query-answer pairs. Implementing our approach requires two key innovations relative to~\cite{2022ip,2023vip}: 1) modifying the IP framework to accommodate large language models (LLMs) for answering queries, and 2) handling ``Unknown'' answers, as illustrated in Figure~\ref{fig:example_intro}. 
Moreover, our method achieves higher F1 score against black-box baselines such as Flan-T5-large and interpretable-by-design methods such as Concept Bottleneck Models. 
Qualitatively, the explanations provided by IP-CRR offer a potential venue for trustworthy and transparent predictions in clinical decision-making.

\section{Related Work}

\myparagraph{Radiology Report Classification} Automatic disease diagnosis from radiology reports is a critical task that enables large-scale use of ML models in clinical settings. Natural language processing methods are widely used to extract diagnostic labels from free-text radiology reports. 
There are two main approaches for this task: rule-based labeling and ML-based labeling. 
Rule-based methods~\cite{chapman2001simple,peng2018negbio,bozkurt2019rule,irvin2019chexpert} rely on predefined rules or regular expressions to extract diseases from the report. However, these methods struggle with the complexity and variability of clinical language. 
ML-based methods~\cite{wang2022radiology,smit2020chexbert,pereira2024automated,holste2024towards} learn the patterns of text automatically without relying on handcrafted rules. 
For example, CheXbert~\cite{smit2020chexbert} is a BERT-based medical domain specialized model pretrained on rule-based annotations, and finetuned on a small expert annotated dataset. 
However, despite their performance, they lack interpretability, making it hard for radiologists to trust their predictions in clinical use.

\myparagraph{Interpretable-by-design Methods}
Interpretable-by-design methods aim to produce explanations that are faithful to the model. State-of-the-art methods ensure this by using an intermediate layer of interpretable factors (e.g., concepts or queries) based on which a prediction is made. For example, Concept Bottleneck Models (CBMs) first map a given input to a list of concept scores, each corresponding to the presence of a concept in the input, and then \textit{linearly} map concept scores to a predicted label. 
CBMs have found many successful applications in domains such as image classification and clinical prediction~\cite{2023lfcbm,2024adacbm,2023lobo,2023probcbm,2023CDM}. However, CBMs are sensitive to having concepts of very high quality and their use of a linear classifier limits their accuracy. In this work, we focus on another framework known as Information Pursuit (IP)~\cite{2022ip,2023vip,covert2023learning}, which provides interpretability via sequentially selecting queries in order of information gain, and then making predictions based on the sequence of query-answer pairs. Therefore, IP aligns more closely with the interactive question-answering process between a doctor and a patient. 
The challenge, however, involves constructing a set of meaningful queries and obtaining their answers. Currently, it has only been explored for image classification, where \cite{2024fmvip} leverages LLMs~\cite{2024gpt} and vision-language models~\cite{li2023blip}. In this work, we aim to leverage natural language inference for text and medical domain. We discuss the framework in detail in Section~\ref{sec:ip-vip}.

\section{Method}

In this section, we describe IP-CRR, the proposed interpretable-by-design model for classifying CRRs. Our pipeline consists of three main components: 1) generate a set of queries from a large number of reports (Sec.~\ref{sec:qry-construct}), 2) design a mechanism to answer queries automatically (Sec.~\ref{sec:qry-construct}), and 3) integrate the Information Pursuit framework (Sec.~\ref{sec:ip-vip}) to conduct interpretable classification.

\begin{figure}[t]
    \centering
    \includegraphics[width=\textwidth]{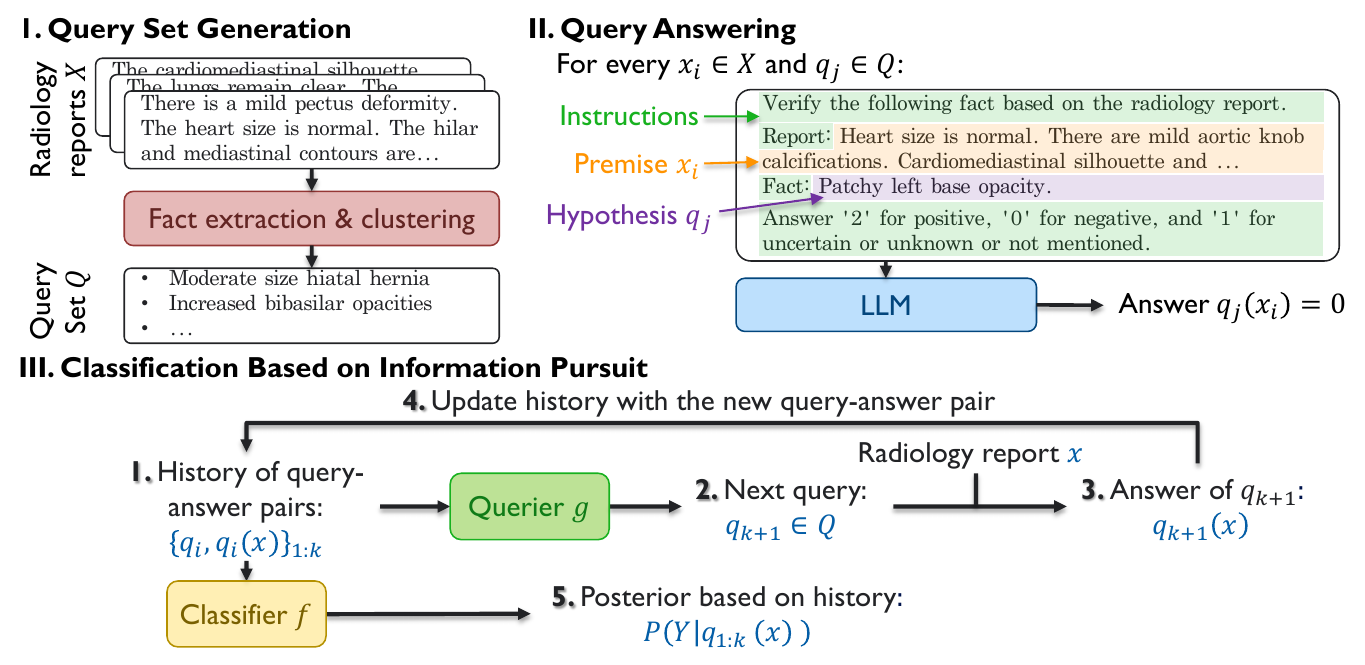}
    \caption{Framework of IP-CRR, which consists of three parts: 1) query set generation by extracting and clustering facts from a large-scale radiology report dataset, 2) query answering based on natural language inference, and 3) interpretable classification based on variational information pursuit.}
    \label{fig:query_set}
\end{figure}

\subsection{Problem Formulation}

We consider the problem of classifying a CRR based on a sequence of interpretable queries. Let $X\in \mathcal{X}$ be a random variable representing a CRR, where $\mathcal{X}$ is the space of all possible CRRs, and $Y\in \mathcal{Y}= \{0, 1\}$ denotes the corresponding binary label. Our goal is to determine $Y$ given $X$ by sequentially querying $X$ with a set $Q$ of predefined, interpretable queries.

Each query $q \in Q$ is a function $q: \mathcal{X} \to \mathcal{A}$, where $\mathcal{A} = \{-1, 0, 1\}$ represents the possible answers: negative, unknown, and positive. 
For example, given a radiology report $x$ and a query $q=$ ``\textit{Is there a decreased opacity on the right?}'', $q(x) = 1$ implies there is a decreased opacity on the right, $q(x)=-1$ implies there is not, and $q(x)=0$ implies the query cannot be answered from the given report. We assume the query set $Q$ is interpretable, task-specific, and sufficient\footnote{That is, $P(Y \mid x) = P(Y \mid \{x' \in \mathcal{X}: q(x) = q(x'), q \in Q\})$. See \cite{2022ip}. }. 

The classification process involves selecting the most informative queries iteratively until sufficient information is obtained to predict $Y$ with high confidence.

\subsection{Constructing and Answering Queries from Existing Databases}\label{sec:qry-construct}

Our final goal is to classify whether a certain disease is present or not in a given CRR. To apply the IP framework, we first need to define a set of interpretable and task-relevant queries about the report. Each query corresponds to a question about whether a specific fact is ``true'', ``false'', or ``unknown'' given the report. For example, the fact ``\textit{normal heart shape}'' can be thought of as the query ``\textit{Is the heart shape normal?}'' While there are many choices for the query set, we argue that curating a sufficiently large set of facts as queries for CRR classification allows us to build interpretable models with good predictive performance. Next, we describe how to construct this query set and answer queries for a CRR.

\myparagraph{Queries $q$} The procedure for obtaining a large set of representative, domain-specific, and interpretable queries is illustrated in Figure~\ref{fig:query_set} Part I. We construct the query set $Q$ by mining facts from a large-scale CRR dataset using a recently developed fact extraction model~\cite{2024cxrfe}. Specifically, we analyze over 220k CRRs from the MIMIC-CXR dataset~\cite{2019mimic} and parse them into more than 660k sentences using the Natural Language Toolkit~\cite{bird-loper-2004-nltk}. Next, we extract over 590k facts from these 660k sentences by first prompting the GPT-3.5 and GPT-4 models to generate facts for a subset of the sentences and then using these sentence-fact pairs as training data to finetune the T5-small~\cite{raffel2020exploring} sequence-to-sequence model for fact extraction. After fact extraction, we encode each fact into a 128-dimension latent space using the fact encoder CXRFE~\cite{2024cxrfe}. We apply $k$-means clustering to all 590k extracted facts in the fact embedding space, then select the nearest fact to each cluster center, and conduct deduplication if any two selected facts have a cosine similarity greater than 0.97. After this process, we obtain 520 representative facts, where each of them is a short fact describing the clinical findings, e.g., ``\textit{moderate size hiatal hernia}'' or ``\textit{increased bibasilar opacities}''.

\myparagraph{Query Answers $q(x)$} Ideally, query answers should be obtained from the report through annotations made by radiologists. However, manual annotation is infeasible due to the large number of reports and queries. Moreover, in contrast to image classification, it often occurs that a query cannot be answered based on the provided radiology report. For example, if the description of the heart is not mentioned in the report, we cannot answer the query ``Is the heart size normal?'' from the report, although the heart is shown in the image. To obtain the query answers in a scalable and accurate manner, we propose to leverage natural language inference (NLI). Specifically, we pass both the query and the CRR as inputs to the Flan-T5-large~\cite{2024flant5} LLM and ask it to predict the query answer as either ``yes'', ``no'' or ``unknown''. An example of the prompt is illustrated in Figure~\ref{fig:query_set} Part II. The prompt consists of three parts: instruction, premise, and hypothesis. The instruction serves as guidance to the model and is fixed for every sample, the premise is the radiology report, and the hypothesis is the fact to be verified as true for the given report. For every report and query, the original model output is 0, 1, or 2 for negative, unknown, or positive. Then we normalize them to form the query answer $q(x) \in \{-1, 0, 1\}$.

\subsection{Classification Based on Variational Information Pursuit}\label{sec:ip-vip}

\myparagraph{IP} 
We aim to classify a report using Information Pursuit (IP), an interpretable-by-design framework that sequentially selects queries about the data in order of information gain. Let $I(q(X);Y)$ denote the mutual information between $q(X)$ and $Y$. Given a sample CRR $x^{\text{obs}} \in \mathcal{X}$, IP selects the queries as follows:
\begin{align}
    q_1 = \argmax_{q \in Q} I(q(X);Y); \quad 
    q_{k+1} = \argmax_{q \in Q} I(q(X); Y \mid q_{1:k}(x^{\text{obs}})).
\end{align}
Here, $q_{k+1} \in Q$ refers to the new query selected at step $k+1$ based on the history of query-answer pairs $q_{1:k}(x^{\text{obs}})$. The algorithm terminates at iteration $L$ when there are no more informative queries, i.e., $\forall q \in Q, I(q(X); Y \mid q_{1:L}(x^{\text{obs}})) = 0$. Furthermore at each iteration $k$, one can also compute the posterior distribution (i.e. the prediction) $P(Y \mid q_{1:k}(x^{\text{obs}}))$. To carry out IP, \cite{2022ip} proposed to learn a VAE to model the joint distribution of all query answers and labels $P(Q(X), Y)$. However, learning a good generative model is challenging when the size of $Q$ is large, so a variational approach is proposed as an alternative.

\myparagraph{V-IP} Rather than modeling the joint distribution of queries and labels to directly compute the most informative queries, Variational Information Pursuit (V-IP)~\cite{2023vip,2024fmvip,covert2023learning} finds a querier $g$ that selects the most informative query $q$ given any random history $S$ of query-answer pairs, and a predictor $f$ that models the posterior distribution of $Y$ given $q$ and $S$, by optimizing the following objective:
\begin{align}\tag{V-IP}
\begin{split}
    \min_{f, g} & \quad \mathbb{E}_{X,S}\left[D_{KL}(P(Y\mid X) \ \| \ \hat{P}(Y\mid q(X), S)) \right] \\ 
    \text{where} & \quad q:= g(S) \in Q \\
    & \quad \hat{P}(Y \mid q(X), S) := f(\{q, q(X)\}\cup S).
    \label{eq:object}
\end{split}
\end{align}

The variational objective aims to minimize the KL-divergence between the true posterior distribution and the posterior distribution given any history and an additional query and its answer. In practice, the classifier and querier are parameterized as neural networks, $f_\theta$ and $g_\eta$, making them efficient for carrying out the IP algorithm and scalable for evaluating large-scale datasets. The history $S$ is represented using a masked vector with length $|Q|$. Since our task is classification, where $Y$ is a categorical variable, the KL-divergence is empirically equivalent to the cross-entropy loss. Once $f_\theta$ and $g_\eta$ are trained, the querier $g_\eta$ serves as a function that directly outputs the query with the highest mutual information~\cite{2023vip}, hence IP can be directly carried out by having $q_{k+1} = g(\{q_{1:k}, q_{1:k}(x)\})$. 

To implement IP-CRR, we first obtain a query set from large amounts of CRRs as described in Sec.~\ref{sec:qry-construct}, then answer each query for each report with an LLM. 
Finally, we apply V-IP to train a querier and a classifier to generate a sequence of interpretable query-answer pairs for the final classification.

\section{Experiments}

In this section, we will demonstrate the effectiveness of IP-CRR at performing CRR classification.

\subsection{Dataset and Implementation Details}

\myparagraph{Dataset} We evaluate the proposed method on a public radiology dataset named MIMIC-CXR~\cite{2019mimic}. In this dataset, each patient is called ``subject'', and each examination is called ``study''. Each study has one radiology report and multi-label ground-truth annotations~\cite{holste2024towards} indicating whether diseases are present. The dataset is randomly split into training, validation, and test sets in a 7:1:2 ratio.

\myparagraph{Implementation Details} 
The architecture of both the querier and the classifier consists of a multilayer perceptron with 5 layers
(see \href{https://github.com/Glourier/MICCAI2025-IP-CRR}{code} for details). 
The IP-CRR model terminates with a confidence threshold of 0.85 or a maximum number of queries of 200. The optimizer is AdamW~\cite{2017adamw} with an initial learning rate of 0.0001 and a cosine annealing decay. The model is trained with a batch size of 128 for 1000 epochs. The loss function is weighted binary cross-entropy to alleviate the class imbalance. All experiments are implemented in PyTorch 2.5.1 using an NVIDIA A5000 GPU with a memory of 24.5GB.

\begin{table}[t]
\centering
\caption{Predictive performance of IP-CRR and baseline models across six tasks, measured by Average precision (AP, left) and F1 score (right). Higher values indicate better performance. The best results among black-box models are \underline{underlined}, while the best interpretable model results are \textbf{bolded}.}

\resizebox{\textwidth}{!}{
\begin{tabular}{l|cccccc|cccccc}
\toprule
\multicolumn{1}{c|}{\multirow{2}{*}{\textbf{Methods}}} & \multicolumn{6}{c|}{\textbf{AP}} & \multicolumn{6}{c}{\textbf{F1}} \\ \cmidrule{2-13} 
\multicolumn{1}{c|}{} & \textbf{LO} & \textbf{CA} & \textbf{SD} & \textbf{CM} & \textbf{PE} & \textbf{PN} & \textbf{LO} & \textbf{CA} & \textbf{SD} & \textbf{CM} & \textbf{PE} & \textbf{PN} \\ \midrule
CXR-BERT (FT-Last) & 0.900 & 0.361 & 0.969 & 0.864 & 0.945 & 0.449 & 0.829 & 0.223 & 0.912 & 0.789 & 0.887 & 0.449 \\
CXR-BERT (FT-All) & \underline{0.984} & \underline{0.992} & \underline{0.970} & \underline{0.964} & \underline{0.962} & \underline{0.641} & \underline{0.987} & \underline{0.991} & \underline{0.978} & \underline{0.982} & \underline{0.953} & \underline{0.541} \\
Flan-T5-large & 0.527 & 0.073 & 0.445 & 0.380 & 0.616 & 0.190 & 0.663 & 0.139 & 0.321 & 0.543 & 0.754 & 0.299 \\
\hline
CBM & 0.947 & 0.345 & 0.934 & 0.791 & 0.874 & 0.432 & 0.884 & 0.241 & 0.853 & 0.738 & 0.801 & 0.431 \\
\textbf{IP-CRR} & \textbf{0.972} & \textbf{0.578} & \textbf{0.959} & \textbf{0.892} & \textbf{0.925} & \textbf{0.468} & \textbf{0.918} & \textbf{0.350} & \textbf{0.889} & \textbf{0.811} & \textbf{0.860} & \textbf{0.451} \\
\bottomrule
\end{tabular}
}
\label{tab:result_both}
\end{table}

\begin{figure}[t]
    \centering
    \begin{minipage}{0.45\textwidth}
        \centering
        \includegraphics[width=\textwidth]{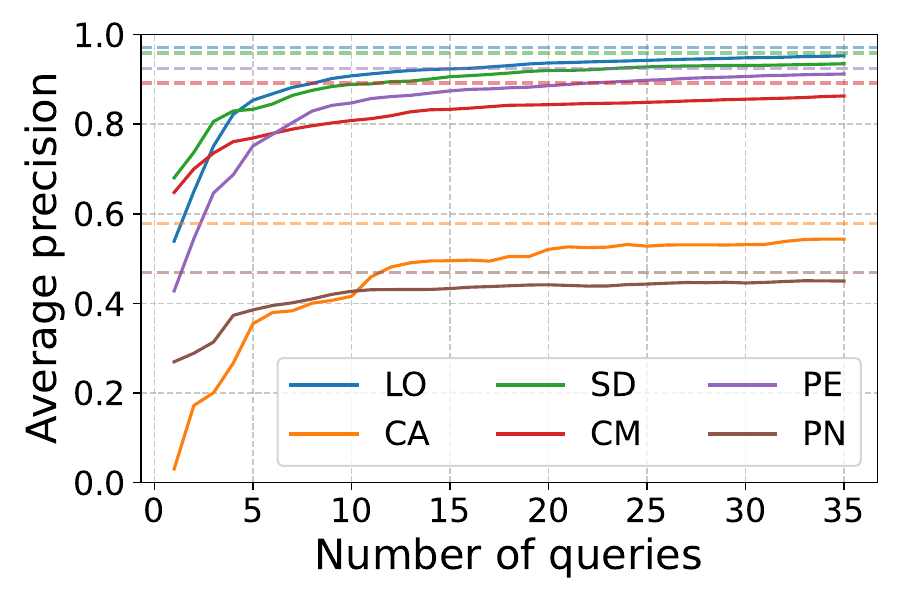}
        \subcaption{Average precision versus number of queries asked. Dashed lines mark the maximum performance over all queries. }
        \label{fig:result_ap}
    \end{minipage}%
    \hspace{0.05\textwidth} 
    \begin{minipage}{0.45\textwidth}
        \centering
        \includegraphics[width=\textwidth]{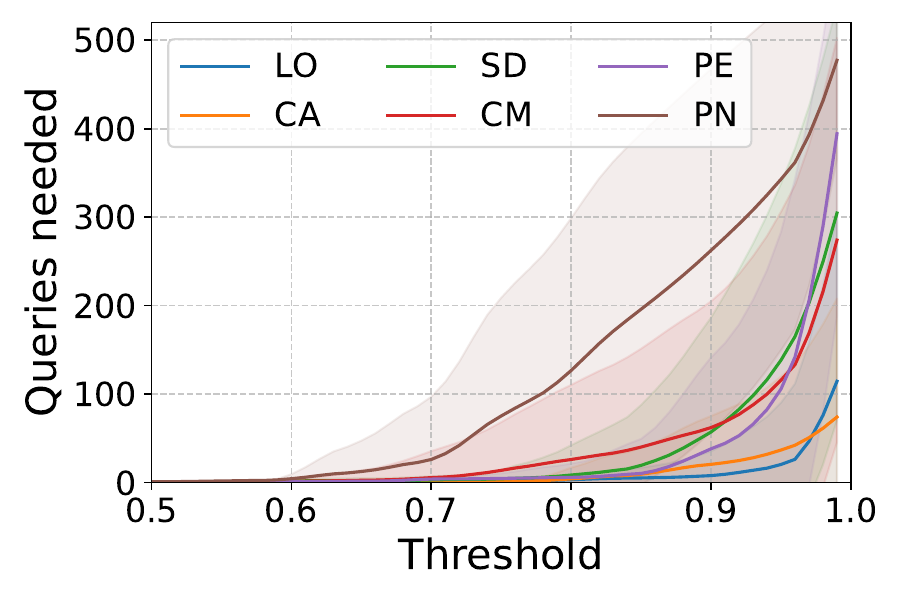}
        \subcaption{Number of queries needed over thresholds on the posterior. Shaded regions indicate the variance across runs. }
        \label{fig:queries_needed}
    \end{minipage}
    \caption{Quantitative evaluation of IP-CRR as the number of queries increases.}
    \label{fig:result}
\end{figure}

\subsection{Quantitative Results and Comparisons}\label{sec:classificaion_result}
We evaluate the performance of IP-CRR with all queries on six binary classification tasks: Lung Opacity (LO), Calcification of the Aorta (CA), Support Devices (SD), Cardiomegaly (CM), Pleural Effusion (PE), and Pneumonia (PN). Baselines include CXR-BERT~\cite{2022cxrbert} finetuned on the last layer (FT-Last) and on all layers (FT-All), Flan-T5-large~\cite{2024flant5} without fine-tuning, and CBM~\cite{2020cbm}. Notably, the first three are black-box models and CBM is interpretable by design. 

As shown in Table~\ref{tab:result_both}, IP-CRR outperforms CXR-BERT (FT-Last), Flan-T5-large, and CBM in average precision (AP) and F1 score on the LO, CA, CM, and PN tasks, and achieves comparable performance on SD. However, compared to CXR-BERT (FT-All), IP-CRR matches its AP on LO, but underperforms on other tasks. We attribute this gap primarily to the CXR-BERT's black-box nature, its more complex BERT-based architecture, and the fact that all its parameters are fine-tuned during training. In contrast, IP-CRR is implemented using a simple MLP with significantly fewer parameters, prioritizing interpretability. 

Furthermore, in Figure~\ref{fig:result_ap}, we evaluate AP as a function of the number of queries. For the LO task, IP-CRR achieves 0.95 AP with just 30 queries, and 0.97 with 100 -- fewer than 20\% of the total of 520 queries. Similar trends hold across other tasks, indicating that IP-CRR generally selects informative queries in its early iterations. Alternatively, we observe that IP-CRR converges to the maximum performance with only a small subset of queries. Compared to the baselines in Table~\ref{tab:result_both}, IP-CRR achieves better performance while using a smaller number of queries than CBMs, which use all available queries. 

Next, in Figure~\ref{fig:queries_needed}, we plot the number of queries needed as the threshold on the posterior changes. The threshold on the posterior approximately measures the confidence the user requires the predictor to reach before terminating. We observe that the IP-CRR predictor requires fewer than 150 queries to achieve a confidence of 0.95 for most of tasks. This demonstrates that IP-CRR is able to find informative queries to make confident predictions.

\subsection{Qualitative Analysis of Query-answer Chains}

\begin{figure}[t]
    \centering
    \includegraphics[width=\textwidth]{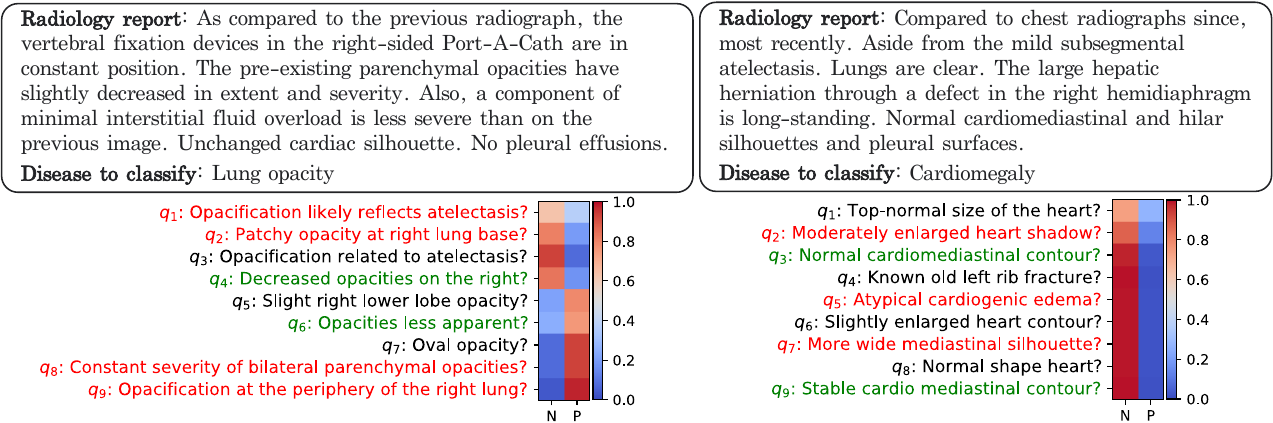}
    \caption{Examples of query-answer chains from IP-CRR. Each row in the colored matrix represents the posterior $P(Y \mid q_{1:k}(x))$ after selecting the query $q_k$. Text in \textcolor{red}{red} means the answers is ``no'', \textcolor{darkgreen}{green} means ``yes'', and black means ``unknown''. ``N'' or ``P'' means negative or positive prediction of the disease, respectively.}
    \label{fig:ip-example}
\end{figure}

The core mechanics of using IP-CRR for interpretable predictions is the iterative process of selecting informative queries. This allows the user to observe how the posterior evolves as the number of queries increases. In Figure~\ref{fig:ip-example}, we demonstrate two examples of query-answer chains obtained using IP-CRR. For the example on the left, our method first selects $q_1=$ ``\textit{Opacification likely reflects atelectasis?}''. From the report, the model infers that $q_1(x)$ is not present in the report. Then, it selects the query $q_2=$ ``\textit{Patchy opacity at right lung base?}''. Continuing in this fashion, after 9 queries, the model concludes that the patient has LO. The same sequential deduction can be observed in the example on the right, highlighting the interpretability IP-CRR offers when applied to CRRs.

\section{Conclusion}
In this work, we have proposed an interpretable-by-design framework for CRR classification. We obtain a large set of queries by mining them from existing CRRs, and answer them using NLI. To evaluate the proposed framework, we conducted experiments on a large-scale CRR dataset and on multiple tasks, with both quantitative and qualitative results that illustrate interpretability and competitive performance. A current limitation is the reliance on a general-domain language model for query answering. We leave the use of stronger medical foundation models for generating more expressive query answers to future work.

\begin{credits}
\subsubsection{\ackname} The authors thank Kaleab A. Kinfu, Uday Kiran Reddy Tadipatri and Prof. Dennis Parra for insightful discussions and feedback during the development of this work. Yuyan, Ryan, and René acknowledge the support from the Penn Engineering Dean's Fellowship, Penn startup funds, and the NIH grant R01NS135613. Pablo was supported in Chile by ANID through iHEALTH (ICN2021\_004), CENIA (FB210017), and the ANID Scholarship Program/Doctorado Becas Chile/2019-21191569.

\subsubsection{\discintname}
The authors have no competing interests to declare that are relevant to the content of this article. 
\end{credits}


\bibliographystyle{splncs04}
\bibliography{References}

\end{document}